%% file: acl_latex.tex
\pdfoutput=1
\documentclass[11pt]{article}

\usepackage[]{acl}

\usepackage{times}
\usepackage{latexsym}
\usepackage[T1]{fontenc}
\usepackage[utf8]{inputenc}
\usepackage{microtype}
\usepackage{inconsolata}
\usepackage{graphicx}
\graphicspath{ {./images/} }

\usepackage{booktabs}
\usepackage{paralist}
\usepackage{tabularx}
\usepackage{pifont}

\newcommand\setalign{4pt}

\title{Seed-Free Synthetic Data Generation Framework\\for Instruction-Tuning LLMs: A Case Study in Thai}

\author{
Parinthapat Pengpun\textsuperscript{‡},
Can Udomcharoenchaikit\textsuperscript{\dag}, \\
\textbf{Weerayut Buaphet}\textsuperscript{\dag}, \textbf{Peerat Limkonchotiwat}\textsuperscript{\dag} \\
\textsuperscript{‡}Bangkok Christian International School, Thailand\\
  \textsuperscript{\dag}School of Information Science and Technology, VISTEC, Thailand \\
  \texttt{parinzee@protonmail.com}
  \\
    \texttt{ \{canu\_pro,weerayut.b\_s20,peerat.l\_s19\}@vistec.ac.th}
  }

\begin{document}
\maketitle
\input{0.abstract}
\input{1.introduction} 
\input{2.related_works} 
\input{3.methodology} 
\input{4.experimental_setting} 

\input{5.experimental_results} 
\input{6.conclusion} 

\bibliography{anthology,custom}

\input{7.appendix}

\end{document}

%% file: 0.abstract.tex
\begin{abstract}
We present a synthetic data approach for instruction-tuning large language models (LLMs) for low-resource languages in a data-efficient manner, specifically focusing on Thai. 
We identify three key properties that contribute to the effectiveness of instruction-tuning datasets: fluency, diversity, and cultural context. 
We propose a seed-data-free framework for generating synthetic instruction-tuning data that incorporates these essential properties. 
Our framework employs an LLM to generate diverse topics, retrieve relevant contexts from Wikipedia, and create instructions for various tasks, such as question answering, summarization, and conversation. 
The experimental results show that our best-performing synthetic dataset, which incorporates all three key properties, achieves competitive performance using only 5,000 instructions when compared to state-of-the-art Thai LLMs trained on hundreds of thousands of instructions. 
Our code and dataset are publicly available at \url{https://github.com/parinzee/seed-free-synthetic-instruct}.
\end{abstract}

%% file: 1.introduction.tex
\section{Introduction}
 
Large Language Models (LLMs) have achieved a near human-level of performance across multitudes of tasks and domains~\cite{openai2024gpt4, geminiteam2024gemini, ma2024eureka, antaki2023capabilities}.
However, many evaluation results have shown that this level of performance is often limited to high-resource languages only, with inconsistent levels of performance for lower-resource languages, i.e., Thai~\cite{xue2024comprehensive, zhang-etal-2023-dont, krause-etal-2023-confidently, huang-etal-2023-languages, ahuja-etal-2023-mega}.
The development of LLMs for low-resource languages is crucial for enabling impactful applications for millions of people worldwide.
Some applications of these LLMs include medical chatbots~\cite{sanna-etal-2024-building}, intelligent tutoring systems~\cite{sonkar-etal-2023-class, afzal-etal-2019-development}, and content moderation tools that could help combat misformation and hate speech~\cite{kumar2024watch}. 
These potential applications have motivated researchers to explore methods for improving LLM performance in low-resource languages.

Recently, researchers developed fine-tuning techniques to improve the performance of LLMs in Thai as well.
SambaLingo~\cite{csaki2024sambalingo} investigated how performing continual pretraining and instruction-tuning on machine-translated English datasets results in good performance in multiple low-resource target languages including Thai.
WangchanX~\cite{phatthiyaphaibun2024wangchanlion} explored using pre-existing Thai datasets to perform instruction-tuning and adapt the SEA-LION model~\cite{sea_lion_2023} to the Thai language.
Typhoon-Instruct~\cite{pipatanakul2023typhoon} adapted LLaMa-3~\cite{llama3modelcard} to the Thai language through continual pretraining on a filtered web corpus, and instruction-tuning on a combination of machine-translated datasets and Thai synthetic datasets generated with Self-Instruct~\cite{wang2023selfinstruct}.
These methods typically use over 50k and sometimes over 100k examples for their instruction-tuning process, making it very costly. Additionally, some of these approaches involve continual pretraining, which further increases the cost and complexity of the model development process.

It remains unclear whether such large datasets are truly necessary for achieving high performance in low-resource languages, as the aforementioned works in Thai LLMs do not address this. However, other works have also shown that LLM alignment in English does not require extensively large datasets~\cite{zhou2023lima, du2023mods}.

Thus, by carefully designing a high-quality synthetic dataset tailored to the target language, we hypothesize that it may be possible to achieve similar performance improvements in Thai while significantly reducing the data requirements and associated costs.


To formulate a high-quality synthetic dataset, we identify three key properties that the datasets used to finetune the current Thai LLMs have:
\begin{compactenum}[\hspace{3pt}(1)]
    \item \textbf{Fluency}: the data is grammatically correct and natural-sounding, enabling the model to learn the proper structure and flow of the language.
    \item \textbf{Diversity}: the data consists of a wide range of topics and domains, allowing it to generalize better to various downstream tasks.
    \item \textbf{Cultural Context}: the data contains instructions and information relating to the culture and beliefs appropriate for the average person from the country of the target language.
\end{compactenum}
These properties are commonly present within the dataset used to train these models. Fluency is inherently present in the way that humans write and thus is within the human-annotated Han Instruct Dataset~\cite{phatthiyaphaibun_2024_10935822} used to train WangchanX. Diversity comes from the fact that existing datasets commonly cover multiple domains. For example, SambaLingo uses a translated version of UltraChat, which covers a wide range of topics. Cultural context is also present in these datasets. For example, Iapp Wiki QA~\cite{kobkrit_viriyayudhakorn_2021_4539916}--- a subset of OpenThaiGPT's training dataset--- includes questions on Thai Wikipedia data. We hypothesize that combining all three properties in a dataset will yield a reasonably performant Thai LLM, even if the dataset is synthetic.

To verify our hypothesis, in this paper, we develop a framework to generate synthetic instruction-tuning datasets with controllable parameters for each of these properties.
We use our framework to create five datasets with varying combinations of the properties as detailed in Section~\ref{sec:training-datasets}.
We then perform instruction-tuning with the base model of LLaMa-3 8B~\cite{llama3modelcard} on each dataset and evaluate their performance on two benchmarks: culture-specific and non-culture-specific datasets.

Our findings suggest that incorporating all three properties in the training data improves the performance of LLMs in low-resource languages, and using only just 5k rows of our dataset for instruction-tuning allows for similar performance against other methods that used 10-100x larger datasets.

We summarize the contribution of our work as follows: 
\begin{compactitem} [\hspace{\setalign}•]
    \item We verify our hypothesis that comparable results to current SOTA Thai LLMs can be achieved by carefully constructing a synthetic dataset that is a fraction of the size of the ones used to train these models.
    \item We propose a seed-data-free framework for synthetically generating finetuning data that is fluent, diverse, and culturally aligned for low-resource languages.
    \item We conduct a large-scale study on data efficiency using 8 LLMs, 5 synthetic datasets, 2 benchmarks, and 7 tasks.
    
\end{compactitem}

%% file: 2.related_works.tex
\section{Related Works}
\subsection{Thai LLMs}
The development of Thai LLMs and other low-resource language LLMs~\cite{csaki2024sambalingo, sea_lion_2023, nguyen2023seallms} have gained attention in the recent year with models such as LLaMa3-8b-WangchanX-sft-Demo~\cite{phatthiyaphaibun2024wangchanlion}, Typhoon-Instruct~\cite{pipatanakul2023typhoon}, and OpenThaiGPT~\cite{OpenThaiGPT2023} being released. 
LLaMa3-8b-WangchanX-sft-Demo leverages a combination of English Datasets: Dolly-15~\cite{DatabricksBlog2023DollyV2}, Math-14k~\cite{hu-etal-2023-adapter}; Human-written Thai datasets (6k); and a Google Gemini~\cite{geminiteam2024gemini} translated versions of Dolly-15k and Math-14k for instruction-tuning, which results in a total of 64k examples. 
Typhoon-Instruct uses both continual pretraining on a filtered subset of Oscar~\cite{ortiz-suarez-etal-2020-monolingual} and finetuned on multiple translated datasets. However, they do not mention the exhaustive list nor the exact number used. 

OpenThaiGPT also performs both continual pretraining and instruction-tuning. Although they do not explicitly mention the dataset composition nor count for the current version (v1.0.0). Previous versions, however, used an extensively large corpus for finetuning consisting of both human-generated and machine-translated data. For example, \verb|openthaigpt-0.1.0-beta| used a combination of 200k samples, and \verb|openthaigpt-gpt2-instructgpt-poc-0.0.1| used 300k samples~\footnote{Information regarding dataset composition is obtained from \href{https://github.com/OpenThaiGPT/openthaigpt}{OpenThaiGPT's Github}}. These previous versions used the GPT-2 architecture~\cite{Radford2019LanguageMA} with 1.5B Parameters. For their latest version, they scaled up the model to LLaMa3-8B and LLaMa3-70B~\footnote{Information regarding architecture obtained from model cards in \href{https://huggingface.co/openthaigpt}{OpenThaiGPT's HuggingFace}}.

While current works in Thai LLM development have focused on scaling the quantity of the dataset and model size, our work aims to improve the performance of Thai LLMs from a data-centric perspective, thereby reducing the costs needed for finetuning a model. Furthermore, our method does not rely on continual pretraining, which also reduces the required computation and time required as well.

\subsection{Synthetic Data Generation for LLMs}
LLMs are dependent on a large number of high-quality datasets in order to achieve good performance~\cite{longpre2023pretrainers}. Traditionally, instruction-tuning datasets were created by human annotators, which is costly and time-consuming. Synthetic dataset generation has emerged as a promising approach to address these limitations.

Self-Instruct~\cite{wang2023selfinstruct} used 175 human-generated instructions as a seed, then prompted GPT to generate unique instructions and tasks, resulting in a dataset consisting of 82k samples. 
%
%
%
%
WizardLM~\cite{xu2023wizardlm} proposed improving LLMs by synthetically generating complex and difficult questions using prompt engineering to increase the difficulty of an instruction or generate a completely new instruction in the same domain as a given instruction. 
In addition, WizardLM uses Alpaca's training data as the initial data and applies the pipeline, which results in a total of 250k samples of instruction-tuning data. 
%
%
UltraChat~\cite{ding-etal-2023-enhancing} proposes a method for generating a large-scale multi-turn dialogue dataset for instruction-tuning. UltraChat obtains context data using various techniques, such as utilizing meta-information from Wikidata and search engines, extracting material types from web pages in the C4 corpus, and prompting GPT-3 to generate instructions for different types of writing tasks. 
This information is then used to perform iterative prompting between two ChatGPT models to simulate user-assistant interactions. 
%
%
Experimental results from these works have demonstrated that synthetic data can improve the performance of LLMs without extensive human effort. 
%

Despite the promising results achieved by existing synthetic data approaches, there remains a gap in the literature regarding the application of these techniques to low-resource languages. Furthermore, these works also utilize high-quality seed instructions, which may be difficult to obtain in low-resource settings. We address this gap by proposing a seed-data-free pipeline for generating instruction-tuning data for low-resource languages.

\subsection{Data Efficient Instruction-Tuning for LLMs}
Training large language models (LLMs) often requires extensive data, posing challenges for low-resource languages due to dataset scarcity and high computational costs. Researchers have explored techniques for efficient instruction-tuning with limited data.

\citet{zhou2023lima} explored constructing a high-quality 1000 sample instruction-tuning data using data from StackExchange, Wikihow, and other online sources. To ensure diversity, the dataset also includes human-annotated instructions as well. In human evaluations, the LIMA model trained on this dataset was found to produce outputs that were strictly preferred to or on par with those from GPT-4 in 43\% of cases, Claude in 46\% of cases, Bard in 58\% of cases, and InstructGPT (DaVinci003) in 65\% of cases. Through ablation studies, they also found that data diversity and quality were more important than quantity for improving the model's performance, as doubling the dataset quantity alone did not contribute to performance increases. Models trained on more diverse data from StackExchange outperformed those trained on a larger quantity of homogeneous data from wikiHow, and models trained on quality-filtered data outperformed those trained on unfiltered data.

\citet{du2023mods} propose a model-oriented data selection (MoDS) approach for efficiently selecting valuable instruction data to fine-tune an LLM. Their method considers instruction quality, coverage, and necessity based on the abilities of the specific target LLM. First, they use a quality evaluation model to filter the original dataset for high-quality instructions. Then, they apply a k-center greedy algorithm to select a maximally diverse seed dataset from this filtered set. The model is initially fine-tuned on this seed data, then further refined with an augmented dataset addressing performance gaps. The final fine-tuning is done on the combination of the seed and augmented data. An LLM fine-tuned with 4,000 MoDS-selected examples outperformed a model trained on the full 214k dataset.

Although these works have shown success in English, there is a lack of literature regarding data-efficient training in low-resource languages. Our framework addresses this gap by providing empirical evidence that using a small but high-quality synthetic dataset can result in competitive performance for an LLM in the Thai language.

%% file: 3.methodology.tex
\section{Synthetic Dataset Generation Pipeline}
\subsection{Overview}
\begin{figure}[h]
    \hspace{-4.5mm}
    \centering
    \includegraphics[scale=0.5]{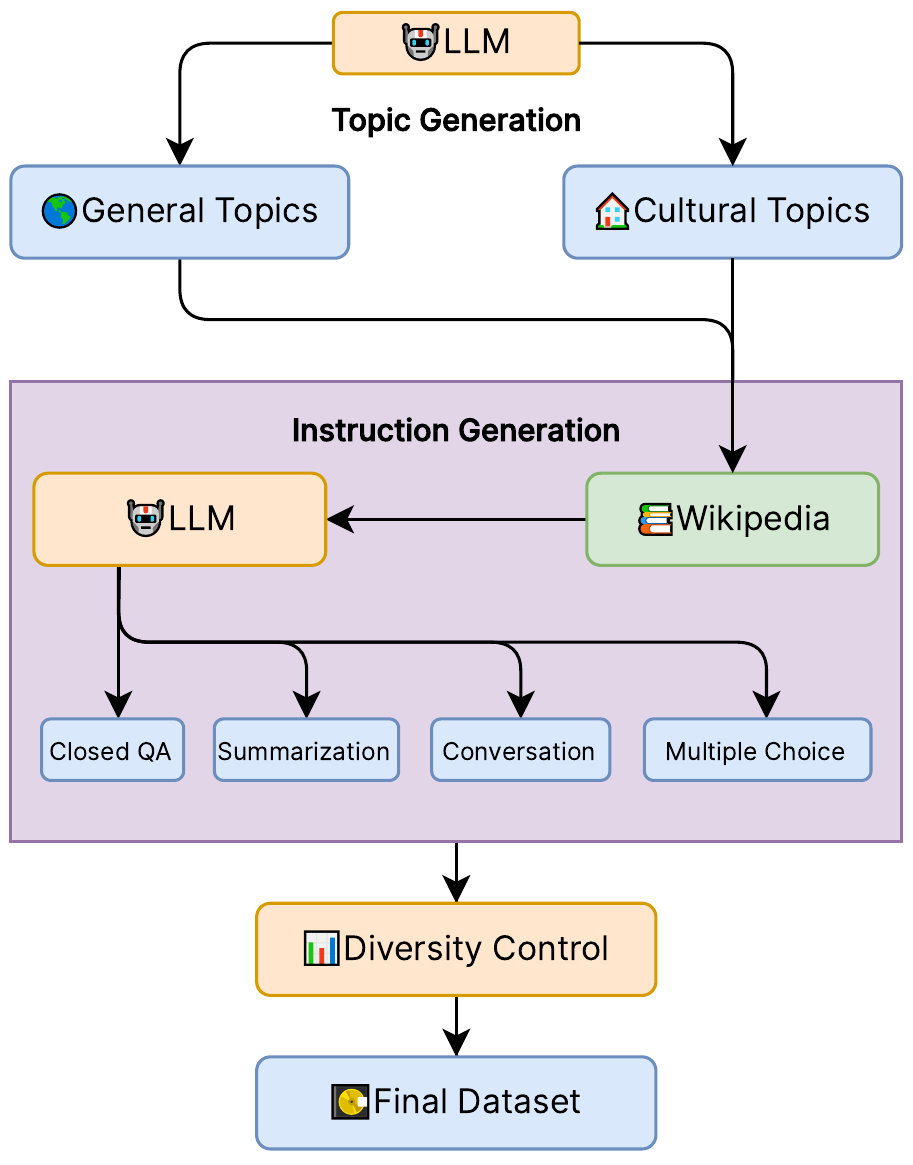}
    \caption{Our proposed framework for generating synthetic instruction-tuning datasets for low-resource languages from scratch with fluency, diversity, and cultural context.}
    \label{fig:framework}
    \vspace{-3.5mm}
\end{figure}

Based on the literature review, current Thai LLMs use extensively large scale datasets. However, we hypothesize that it may be possible to create an LLM model that is comparable to existing Thai LLMs while only using a fraction of the SFT data for finetuning. To verify our hypothesis, we construct and train on 5 synthetic datasets with varying combinations of the three aforementioned properties: Fluency, Diversity, and Cultural Context.
As shown in Figure~\ref{fig:framework}, our synthetic datasets are generated through our framework as follows. The pipeline uses an LLM, in this case, Claude-3 Haiku, to first randomly generate a given number of topics that either are general topics or relate to a specific culture. Using the topics, we search Wikipedia for a related text and then prompt Haiku to generate instructions related to that text. 
Our pipeline generates instructions in the target language (Thai) directly for 4 tasks: Closed Question Answering (Closed QA), Summarization, Conversation, and Multiple Choice.
The data then goes through a diversity control step, where we filter out closely related samples using their semantic embedding vectors to ensure a high-diversity dataset.
We perturb the configuration of the pipeline to obtain the 5 synthetic datasets.

\subsection{Model Selection}
We choose Claude-3 Haiku as the LLM for our synthetic data generation pipeline for several reasons.
First, it has demonstrated strong performance across various natural language tasks~\cite{anthropic2024claude3}, making it well-suited for generating high-quality instruction data.
Second, it is relatively low cost for model of its performance when compared to other state-of-the-art models, this allows for a more cost-efficient dataset generation process.
Third, from our observation, Claude-3 Haiku produced Thai output that is much more fluent and coherent than other LLMs in a similar price range, such as GPT-3.5-Turbo.
This aligns with \citet{enis2024llm}, which has shown that Claude-3 is a strong translator, indicating a good multilingual understanding.
Claude-3's tokenizer is also more efficient in tokenizing Thai characters when compared to GPT-3.5's tokenizer (Claude's tokenizer uses fewer tokens for Thai text).

\subsection{Topic Generation}
We separate our categorization of topics into 2 categories: General Topics and Cultural Topics. For both of these categories, we use a temperature of 0.95. We then prompt Haiku to randomly generate these topics; the specific prompts are below. We repeat this process until we obtain the desired amount of topics. Afterward, the topics are filtered for duplicates and removed.

\noindent
\textbf{General Topics Prompt:} 
\begin{quote}
\emph{Please generate 20 completely random topics. These can be about absolutely anything from everyday conversation, advice, random thoughts, mathematics, science, history, philosophy, etc. Each topic should be a short phrase or sentence. Ensure your output is in the format of a list of strings, where each string is a topic. Your output should be one line in the aforementioned format without anything else.} 
\end{quote}

\noindent
\textbf{Cultural Topics Prompt:} 
\begin{quote}
\emph{You are a native Thai person with expert knowledge of Thai culture, history, language, and customs. Ensure that everything you act, do, say, and generate matches with this fact. Please generate 20 completely random topics relating to your culture. These can be about anything related to your culture such traditions, history, food, language, etc. Each topic should be...}
\end{quote}
The rest of this prompt is omitted as it is the same as the General Topics Prompt.

\subsection{Instruction Generation}
\textbf{Context Selection/Generation} Given a topic, the pipeline first randomly chooses whether to select a related context from Wikipedia or not. If a random context from Wikipedia is not chosen, we prompt Haiku to generate a context related to the topic based on a randomly selected style from this list: news article, blog post, text messages, fictional short story, video transcript, song, poem, scientific study, medical report, social media post with replies, email, tweet, or a how-to article. Otherwise, we search Wikipedia through its MediaWiki API using the topic as our query. We find the top 10 most similar articles to the topic and randomly pick one. We split each article based on their sections, and each of those serves as one context for the instruction.

The following describes our goal and hyperparameters in prompting Haiku to generate instructions for each of these tasks. The full prompt for each task is described in Appendix~\ref{appendix:prompts_for_task}.

\noindent
\textbf{Closed Question Answering.} For this task, we take the context from the previous step and prompt Haiku to generate 5 question-answer pairs for each context. We emphasize our prompting to ensure that Haiku generates questions that come from roughly throughout the whole context. Furthermore, we also noticed that sometimes, Haiku would use ``common knowledge'' that is assumed to be known when generating answers to its questions. To alleviate this, we also emphasize not using any ``external information'' in our prompt. We use a temperature of 0.35 for this task.

\noindent
\textbf{Summarization.} We take the context from the previous step and prompt Haiku to generate a summary for context. The summary is generated in one of three styles, which is randomly picked and embedded into the prompt: bullet points, paragraphs, or numbered lists. We use a temperature of 0.35 for this task.

\noindent
\textbf{Conversation.} This task does not require any context and is designed to mimic how a human might talk to a chatbot— hence the name of the task is called ``Conversation.'' We prompt Haiku to generate a random conversation between an AI assistant and human that relates to a given topic. We emphasize that the assistant must maintain a friendly and casual conversation. We use a temperature of 0.8 for this task.

\noindent
\textbf{Multiple Choice.} We use the context from the previous step and prompt Haiku to generate a question regarding the context and possible answer choices (with only 1 correct answer). Please note that we later also shuffled the answer choices as we noticed that Haiku has a tendency to put correct answer choices as the first one. Because we later do this, we also prompt Haiku to not use any ordinal information in the answer choices, i.e., ``the first and third choice'' or ``B and D''. We use a temperature of 0.4 for this task.

\subsection{Diversity Control}
Although we use relatively high temperatures for these tasks, there may still be cases where we get multiple samples of instructions that are quite similar. To ensure that our dataset is diverse, we filter out any samples that are closely related to each other semantically. We first use BGE-M3~\cite{bge-m3} to encode all of the samples. BGE-M3 is chosen due to its exceptional Thai performance. The samples are formatted by concatenating the instruction, context, and output. For each sample, we do an approximate nearest neighbor search across the whole dataset. If the cosine similarity of the nearest match of that sample is over 0.95, we remove that sample. This process ensures that our final dataset is sufficiently diverse.

%% file: 4.experimental_setting.tex
\section{Experimental Setting}
\subsection{Training Datasets}
\label{sec:training-datasets}
We generate 5 synthetic datasets with varying combinations of fluency, diversity, and cultural contexts using our pipeline to demonstrate that all 3 properties are required for a high-performance model. Each of these datasets has 5,000 samples of instructions. This number is similar to other works in other languages (e.g., LIMA~\cite{zhou2023lima} used 1,000 samples, and MoDS~\cite{du2023mods} used 4,000 samples).

\begin{compactitem}[\hspace{\setalign}•]
\item \textbf{Fluency + Cultural Context + Diversity (F+C+D+):} Constructed by running the pipeline fully with quality control using 750 randomly generated topics in total (400 cultural and 300 general). This dataset is generated in Thai directly by our pipeline and is not translated.
\item \textbf{Fluency Only:} Constructed by running the pipeline with only 10 randomly generated topics (general topics only) and without any diversity control to reduce overall diversity. Only general topics were used to ensure no cultural context.
\item \textbf{Diversity Only:} Constructed by running the pipeline with diversity control using 750 general topics. To artificially reduce fluency, we use nllb-200-distilled-600M~\cite{nllbteam2022language} to translate all samples to English and back-translate them to Thai again. This effectively simulates using machine translation to translate an English dataset to Thai. This dataset is constructed to demonstrate the impacts of not having fluency or cultural context.
\item \textbf{Cultural Context Only:} Constructed by using the F+ C+ D+ dataset as a basis. We randomly select 1000 samples; then, we use NLLB to translate them into English. We then use QCPG~\cite{bandel-etal-2022-quality} to paraphrase the dataset. For each sample, we perform 4 paraphrases, resulting in a total count of 5,000 (4,000 paraphrases + 1,000 originals), thereby reducing the overall domain diversity. Then, we translate everything back to Thai again, reducing fluency.
\item \textbf{No Properties:} We randomly select 1,000 rows from the UltraChat-200k dataset (no Thai cultural context). We use QCPG to perform paraphrasing— generating 4 paraphrases for each sample (reduce diversity), resulting in a total count of 5,000. Then, we translate everything to Thai using NLLB (reduce fluency).
\end{compactitem}

\subsection{Models}
We perform instruction finetuning of the base version of Llama-3 8B using these datasets with QLoRa on a single RTX 3090. The total amount of GPU hours used is around 80 hours. Hyperparameters are shown in Table~\ref{table:hyper-params}. In addition to our own models, we also evaluate standard Thai LLMs, such as Typhoon-Instruct-v1.5 8B, OpenThaiGPT-v1.0.0 8B, and LLaMa3-8b-WangchanX-sft-Demo.

\begin{table}[h!t]
    \small
    \centering
        \begin{tabular}{ll}
        \hline
        \textbf{Hyperparameter} & \textbf{Value} \\
        \hline
        Load in 4-bit & True \\
        Sequence Length & 4000 \\
        Adapter Type & QLoRA \\
        LoRA Rank & 32 \\
        LoRA Alpha & 16 \\
        LoRA Dropout & 0.05 \\
        LoRA Target Linear & True \\
        Grad. Accum. Steps & 8 \\
        Micro Batch Size & 1 \\
        Number of Epochs & 3 \\
        Optimizer & Paged AdamW 8bit \\
        Learning Rate & 0.00015 \\
        BF16 Precision & True \\
        Grad. Checkpointing & True \\
        Flash Attention & True \\
        Warmup Ratio & 0.5 \\
        Evals per Epoch & 1 \\
        Saves per Epoch & 1 \\
        Weight Decay & 0.0 \\
        Seed & 42 \\
        \hline
        \end{tabular}
    \caption{Hyperparameters used to finetune our models.}
    \vspace{-5mm}
    \label{table:hyper-params}
\end{table}

\subsection{Evaluation}
We use WangchanThaiInstruct~\footnote{\url{https://huggingface.co/datasets/airesearch/WangchanThaiInstruct_7.24}} as our benchmark as it provides both a Thai \emph{culture-specific} version and a \emph{non-culture-specific} version. It consists of 6,287 samples in total (both versions combined) spanning 3 domains: Legal, Medical, and Finance. The dataset is created and quality assured by human annotators during the whole process. Using this dataset allows us to assess the performance of our LLMs on instructions that require an understanding of Thai culture, as well as instructions that are more general in nature. 

Both versions of the benchmark consist of seven tasks in total: \textbf{Brainstorming}, which evaluates the model's ability to generate creative ideas and solutions based on a given prompt or scenario; \textbf{Classification}, which requires the model to assign a given input to one or more predefined categories; \textbf{Closed QA}, where the model must locate relevant information within a given text to answer a question; \textbf{Creative Writing}, which assesses the model's ability to generate coherent, engaging, and creative pieces of writing based on a prompt or theme; \textbf{Open QA}, similar to Closed QA but with more open-ended questions that may not have a single, definitive answer within the provided text; \textbf{Multiple Choice}, where the model must select the most appropriate or correct answer from a set of options; and \textbf{Summarization}, which involves generating a concise and coherent summary of a given piece of text. 
By evaluating the Thai LLMs on these diverse tasks and domains, we can gain a comprehensive understanding of their performance across different aspects of language understanding and generation.

\begin{table*}[ht!]
\centering
\small
\caption{Average evaluation results across all 7 tasks on the Thai Culture and General Test Sets. F, C, and D denote Fluency, Culture, and Diversity, respectively. The plus sign (+) indicates the presence of the corresponding attribute, while the minus sign (-) indicates its absence.}
\begin{tabular}{lrrrrrrrr}
\hline
\textbf{Metric} & \textbf{F- C- D-} & \textbf{F+ C- D-} & \textbf{F- C+ D-} & \textbf{F- C- D+} & \textbf{F+ C+ D+} & \textbf{WangchanX} & \textbf{Typhoon} & \textbf{OpenThai} \\
\midrule
BERTScore & 45.90 & 57.30 & 49.60 & 48.20 & \underline{69.50} & 68.80 & \textbf{74.10} & 64.50 \\
BLEU & 0.02 & 0.01 & 0.00 & 0.00 & 0.10 & \underline{2.24} & \textbf{2.32} & 0.95 \\
ChrF & 4.38 & 5.18 & 2.90 & 2.74 & 9.47 & \textbf{17.28} & \underline{17.21} & 14.54 \\
METEOR & 2.20 & 3.70 & 1.70 & 1.70 & 6.70 & \underline{11.30} & \textbf{12.70} & 8.20 \\
ROUGE-1 & 1.30 & 3.60 & 1.80 & 1.90 & 7.70 & \underline{13.40} & \textbf{20.70} & 12.20 \\
ROUGE-2 & 0.20 & 1.00 & 0.20 & 0.30 & 3.30 & \underline{5.80} & \textbf{11.80} & 5.60 \\
ROUGE-L & 1.20 & 3.60 & 1.80 & 1.90 & 7.50 & \underline{12.60} & \textbf{20.00} & 11.70 \\
ROUGE-Lsum & 1.20 & 3.50 & 1.80 & 1.90 & 7.60 & \underline{12.70} & \textbf{20.00} & 11.60 \\
SQuAD F1 & 0.50 & 2.80 & 0.82 & 0.82 & 5.32 & \textbf{8.30} & \underline{7.10} & 3.58 \\
\hline
\multicolumn{9}{l}{\textit{(Thai Culture Test Set)}}
\vspace{0.5em}
\end{tabular}
\begin{tabular}{lrrrrrrrr}
\hline
\textbf{Metric} & \textbf{F- C- D-} & \textbf{F+ C- D-} & \textbf{F- C+ D-} & \textbf{F- C- D+} & \textbf{F+ C+ D+} & \textbf{WangchanX} & \textbf{Typhoon} & \textbf{OpenThai} \\
\midrule
BERTScore & 50.90 & 59.70 & 52.60 & 49.50 & \underline{73.20} & 72.20 & \textbf{76.50} & 67.50 \\
BLEU & 0.04 & 0.01 & 0.00 & 0.00 & 0.08 & \underline{2.08} & \textbf{2.61} & 0.88 \\
ChrF & 4.91 & 5.04 & 2.82 & 2.56 & 9.56 & \underline{17.24} & \textbf{17.53} & 14.84 \\
METEOR & 2.50 & 3.50 & 1.70 & 1.60 & 6.70 & \underline{11.10} & \textbf{12.90} & 8.40 \\
ROUGE-1 & 1.50 & 3.00 & 1.50 & 1.60 & 6.60 & \underline{14.00} & \textbf{18.80} & 10.40 \\
ROUGE-2 & 0.40 & 1.10 & 0.20 & 0.30 & 2.70 & \underline{7.10} & \textbf{10.00} & 4.40 \\
ROUGE-L & 1.50 & 3.00 & 1.50 & 1.60 & 6.40 & \underline{13.30} & \textbf{18.10} & 9.90 \\
ROUGE-Lsum & 1.50 & 3.00 & 1.50 & 1.60 & 6.50 & \underline{13.40} & \textbf{18.00} & 9.90 \\
SQuAD F1 & 0.86 & 2.29 & 0.83 & 0.73 & 4.58 & \textbf{7.43} & \underline{7.26} & 3.31 \\
\hline
\multicolumn{9}{l}{\textit{(General Test Set)}}
\end{tabular}
\label{tab:results}
\end{table*}

\noindent
\textbf{Metrics.} We follow the WangchanX-10k's suggested metrics for evaluation. We use BLEU, METEOR, ChrF, ROUGE, and BERTScore to measure the performance in these tasks. However, we do note that the WangchanX-10k mentioned that BERTScore is the most reliable metric as it measures semantic similarity, while other traditional metrics yield inconclusive results.

%% file: 5.experimental_results.tex
\section{Experimental Results}

\subsection{Main Results}
\textbf{Results.} Table~\ref{tab:results} presents the average evaluation results across all tasks on both the Thai Culture Test Set and the General Test Set. Our synthetic datasets are denoted by the presence (+) or absence (-) of three key attributes: Fluency (F), Culture (C), and Diversity (D). The best-performing model for each metric is highlighted in bold, while the second-best model is underlined. On the Thai Culture Test Set, our best-performing synthetic dataset, F+ C+ D+, which incorporates all three key attributes, achieves the second-highest BERTScore of 69.50\%, surpassing WangchanX (68.80\%) and OpenThaiGPT (64.50\%). The Typhoon-Instruct model obtains the highest BERTScore of 74.1\%. The results on the General Test Set follow a similar pattern, with F+ C+ D+ maintaining its second-place position in terms of BERTScore 73.20\%, outperforming WangchanX 72.2\% and OpenThaiGPT 67.50\%. The Typhoon-Instruct model achieves the highest BERTScore of 76.5\%. The full evaluation results for each task are listed in Appendix~\ref{appendix:full_eval_results}.

\textbf{Discussion.} The experimental results demonstrate the effectiveness of our data-centric approach for improving the performance of Thai LLMs, particularly when considering the BERTScore metric, which is deemed the most reliable by the benchmark authors. F+ C+ D+ achieves the second-highest BERTScore on both the Thai Culture Test Set and the General Test Set, surpassing WangchanX and OpenThaiGPT, suggesting that our data generation pipeline is capable of producing high-quality data that can enhance the model's performance on a wide range of tasks, while being still data-efficient. Namely, we use only 5,000 samples of synthetic data to surpass OpenThaiGPT (200k samples + pretraining) and WangchanX (64k samples), both of which use a mix of human-annotated and machine-translated data.

The consistent top performance of F+ C+ D+ and the lower performances of other synthetic sets, which only consist of one property, demonstrates that all three properties are required to build a strong synthetic dataset. In conclusion, all these results verify our hypothesis that it is indeed possible to construct a small synthetic dataset that performs competitively against much larger datasets.

\subsection{Error Analysis}
\begin{figure}[h]
    \centering
    \includegraphics[scale=0.625]{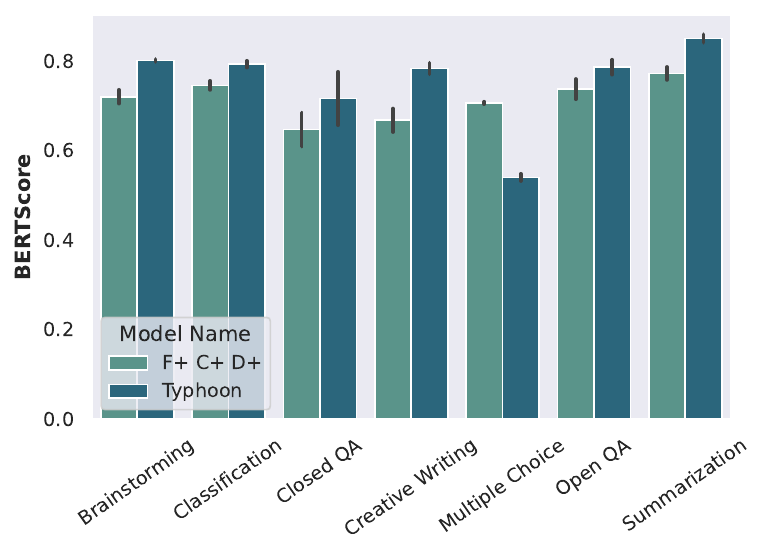}
    \caption{Comparison of BERTScores of our best synthetic model and Typhoon-Instruct on the average scores from both test sets. We also performed Wilcoxon rank-sum tests~\cite{c4091bd3-d888-3152-8886-c284bf66a93a} comparing F+ C+ D+ against Typhoon-Instruct for each task on both the Thai culture-specific and general test sets, and found that the differences were statistically significant (p < 0.05) for all tasks, with an average Wilcoxon statistic of -6.512 and an average p-value of 0.00073 across all comparisons.}
    \label{fig:head_to_head}
    \vspace{-3.5mm}
\end{figure}

In this study, we demonstrate error analysis across different tasks to decipher why our model performs worse than current Thai LLMs in certain tasks. As evidenced in Figure~\ref{fig:head_to_head}, our model performs slightly lower than Typhoon-Instruct in some tasks. When we examine these tasks, it is evident that the tasks with the largest gaps are Brainstorming, Creative Writing, and Summarization.

\begin{figure}[ht]
    \centering
    \includegraphics[scale=0.625]{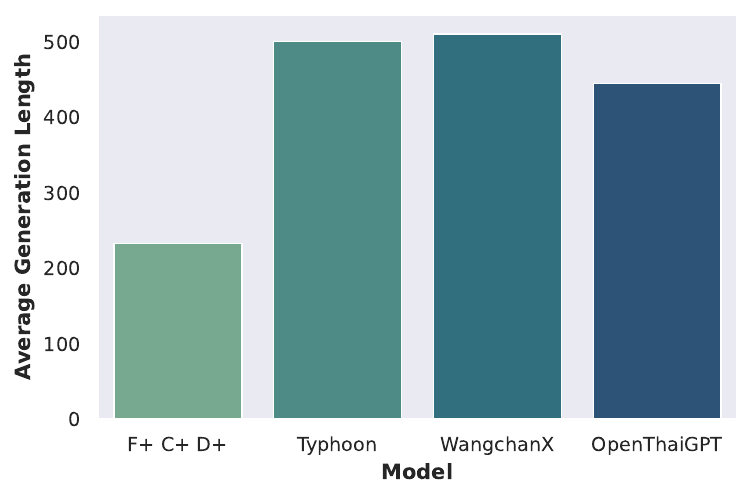}
    \caption{Comparison of average generation lengths across all tasks and both benchmarks. A Wilcoxon rank-sum test was conducted to compare the generation lengths of our best model (F+ C+ D+) and Typhoon-Instruct. The results showed a statistically significant difference (W = -54.233, p < 0.00001), indicating that our model generates significantly shorter outputs compared to Typhoon-Instruct.}
    \label{fig:avg_gen_len}
    \vspace{-3.5mm}
\end{figure}

After investigation, we discovered that our model has a tendency to produce shorter and more concise responses on average. This is shown in Figure~\ref{fig:avg_gen_len}. This could lead to it omitting some information that the reference includes. Hence this leads to a lower score on these open-ended tasks. Since this does not impact tasks that require short responses (i.e., Classification and Open QA), we can see that the difference between Typhoon-Instruct and F+ C+ D+ is much smaller. Furthermore, our model even beats Typhoon-Instruct in Multiple Choice by a large margin. The fact that our model produces shorter responses on average also explains why our model has lower scores when using evaluated n-gram based metrics, which effectively measure text overlap. We conjecture that our model's tendency for brevity stems from the fact that our synthetic data pipeline currently only generates short single-turn dialogues. However, other Thai LLMs, such as Typhoon-Instruct, are trained on machine-translated versions of long multi-turn dialogue datasets like UltraChat. 

%% file: 6.conclusion.tex
\section{Conclusion and Future Work}
In conclusion, this study demonstrates the effectiveness of a data-centric approach for improving the performance of large language models in Thai, a low-resource language. 
We identified three key properties that lead to well-performing Thai LLMs: fluency, diversity, and cultural context. 
We proposed a seed-data-free framework for generating high-quality instruction-tuning data that incorporates these properties. 
Experiments conducted across multiple models, synthetic datasets, benchmarks, and tasks provide empirical evidence that it is possible to achieve competitive results compared to state-of-the-art Thai LLMs trained on 10-100x larger datasets. 
%
%
While our model tends to generate more concise responses compared to the top-performing Typhoon-Instruct model, impacting its performance on open-ended generative tasks, it still achieves impressive results overall, beating other models such as OpenThaiGPT-v1.0.0 and achieving comparable results to WangChanX LlaMa3-8B SFT Demo. 
%

For future work, there are several promising directions to explore. 
One important avenue is to extend our pipeline to generate multi-turn dialogue datasets, which can help alleviate the length issues observed in the current study and enhance the model's ability to handle more realistic, conversation-based scenarios. 
Additionally, conducting experiments with a stronger base model, such as upgrading from Claude-3 Haiku to a more advanced model, could potentially yield even better performance without requiring significant modifications to the data generation process. 
To assess the generalizability of our approach, it would be valuable to expand our experiments to other low-resource languages, adapting the framework to handle different linguistic properties and cultural contexts. 

\section*{Acknowledgements}
We extend our sincere gratitude to Potsawee Manakul for his invaluable assistance during the early stages of this project. 

This research has received funding support from the NSRF via the Program Management Unit for Human Resources \& Institutional Development, Research and Innovation Grant Number B46G670083.

\section*{Limitations}
Our limitation in this paper is we did not investigate the optimal combination of synthetic and human-generated data that could provide insights into the most effective data composition strategies. 
This could involve comparing the performance of models trained solely on synthetic data with those trained on a combination of synthetic and carefully filtered human-generated data. 
In addition, conducting extensive human evaluations would be crucial for assessing the practical usability and perceived quality of the generative models.

%% file: 7.appendix.tex
\clearpage

\appendix
\section{Prompts for Each Task in Instruction Generation}
\label{appendix:prompts_for_task}
\textbf{Closed Question Answering:} \begin{quote}
\emph{Generate 5 questions focusing on different aspects / parts of this given context. Use only the given context to create your questions. Do not use external information. <context>{[}context{]}</context> Ensure your output is in the format of a list of dictionaries, where each dictionary contains a `question' key and an `answer' key. Your output should be one line in the aforementioned format without anything else.}
\end{quote}

\textbf{Summarization:} \begin{quote}
\emph{Generate a concise summary in [summary style] format of the following context related to [topic]: <context> [context] </context> Ensure your output is in the format of a dictionary with a `summary' and `instruction' key, where `summary' is your summary in the specified format and 'instruction' is a sentence you would instruct someone to get this summary (for example: `Please summarize in [summary style] format the following text passage'). Your output should be one line in the aforementioned format, and in the correct language without anything else.}
\end{quote}

\textbf{Conversation:} \begin{quote}
\emph{Generate a conversation between a user and an AI assistant on the topic of [topic]. The user's message should be a question or a statement related to [topic], and the AI assistant should provide a relevant, engaging response to maintain a friendly and casual conversation. The output should be in the following format: <format>Input: User's message Output: AI assistant's response</format> Ensure your output contains ONLY ONE input-output pair exactly in the specified format without any additional text.}
\end{quote}

\textbf{Multiple Choice:} \begin{quote}
\emph{Generate a multiple-choice question focusing on the given context. The question should only have one correct choice. Use only the given context to create your question and answer choices. Do not use external information. <context>[context]</context> DO NOT USE any ordinal information (DO NOT USE eg: first answer is correct, all of the above is correct, etc) of the choices to answer your question as the choices will be shuffled later. Ensure your output is in the following format:<format> Question: Your question Choices: - [Choice 1] - [Choice 2] - [Choice 3] - [Choice 4] Answer: [Explaination + Reasoning + Correct Answer (in this order exactly)] </format> Your output should contain ONLY ONE multiple-choice question exactly in the specified format without any additional text.}
\end{quote}

\section{Full Evaluation Results For Every Task}
\label{appendix:full_eval_results}
\begin{compactitem}
    \item \textbf{Brainstorming}: Table~\ref{tab:brainstorming_results}
    \item \textbf{Classification}: Table~\ref{tab:classification_results}
    \item \textbf{Closed Question Answering}: Table~\ref{tab:closedqa_results}
    \item \textbf{Creative Writing}: Table~\ref{tab:creativewriting_results}
    \item \textbf{Multiple Choice}: Table~\ref{tab:multiplechoice_results}
    \item \textbf{Open Question Answering}: Table~\ref{tab:openqa_results}
    \item \textbf{Summarization}: Table~\ref{tab:summarization_results}
\end{compactitem}

\begin{table*}[ht!]
\centering
\small
\caption{Average evaluation results for the Brainstorming task on the Thai Culture and General Test Sets. F, C, and D denote Fluency, Culture, and Diversity, respectively. The plus sign (+) indicates the presence of the corresponding attribute, while the minus sign (-) indicates its absence.}
\begin{tabular}{lrrrrrrrr}
\hline
\textbf{Metric} & \textbf{F- C- D-} & \textbf{F+ C- D-} & \textbf{F- C+ D-} & \textbf{F- C- D+} & \textbf{F+ C+ D+} & \textbf{WangchanX} & \textbf{Typhoon} & \textbf{OpenThai} \\
\midrule
BERTScore & 54.63 & 55.38 & 51.58 & 53.41 & 70.48 & 68.99 & \textbf{80.43} & \underline{71.76} \\
BLEU & 0.00 & 0.00 & 0.00 & 0.00 & 0.04 & 0.39 & \underline{1.20} & \textbf{2.13} \\
ChrF & 5.03 & 3.92 & 3.04 & 3.52 & 9.54 & 14.51 & \underline{18.41} & \textbf{22.42} \\
METEOR & 2.59 & 2.04 & 1.54 & 1.80 & 5.68 & 7.71 & \underline{10.98} & \textbf{11.43} \\
ROUGE-1 & 2.73 & 1.84 & 1.55 & 0.95 & 6.20 & 12.03 & \underline{21.96} & \textbf{22.01} \\
ROUGE-2 & 0.48 & 0.11 & 0.00 & 0.00 & 2.77 & 4.45 & \textbf{11.42} & \underline{11.13} \\
ROUGE-L & 2.84 & 1.80 & 1.56 & 0.96 & 6.04 & 11.41 & \underline{20.78} & \textbf{21.45} \\
ROUGE-Lsum & 2.82 & 1.87 & 1.52 & 1.01 & 6.00 & 11.44 & \underline{20.58} & \textbf{21.59} \\
\hline
\multicolumn{9}{l}{\textit{(Thai Culture Test Set)}}
\vspace{0.5em}
\end{tabular}

\begin{tabular}{lrrrrrrrr}
\hline
\textbf{Metric} & \textbf{F- C- D-} & \textbf{F+ C- D-} & \textbf{F- C+ D-} & \textbf{F- C- D+} & \textbf{F+ C+ D+} & \textbf{WangchanX} & \textbf{Typhoon} & \textbf{OpenThai} \\
\midrule
BERTScore & 55.80 & 58.46 & 55.55 & 58.54 & \underline{73.55} & 71.76 & \textbf{79.79} & 71.64 \\
BLEU & 0.00 & 0.00 & 0.00 & 0.00 & 0.02 & 0.50 & \underline{1.11} & \textbf{1.33} \\
ChrF & 5.14 & 4.27 & 3.15 & 3.94 & 9.29 & 16.38 & \underline{17.89} & \textbf{19.85} \\
METEOR & 2.65 & 2.45 & 1.75 & 2.28 & 5.60 & 8.56 & \textbf{11.26} & \underline{10.74} \\
ROUGE-1 & 1.86 & 1.91 & 0.75 & 2.06 & 4.66 & 15.37 & \textbf{21.08} & \underline{17.72} \\
ROUGE-2 & 0.21 & 0.30 & 0.10 & 0.52 & 1.59 & 7.19 & \textbf{11.35} & \underline{8.42} \\
ROUGE-L & 1.82 & 1.91 & 0.75 & 1.99 & 4.54 & 14.80 & \textbf{19.93} & \underline{16.67} \\
ROUGE-Lsum & 1.81 & 1.92 & 0.74 & 2.00 & 4.52 & 14.84 & \textbf{20.02} & \underline{16.72} \\
\hline
\multicolumn{9}{l}{\textit{(General Test Set)}}
\end{tabular}
\label{tab:brainstorming_results}
\end{table*}

\begin{table*}[ht!]
\centering
\small
\caption{Average evaluation results for the Classification task on the Thai Culture and General Test Sets. F, C, and D denote Fluency, Culture, and Diversity, respectively. The plus sign (+) indicates the presence of the corresponding attribute, while the minus sign (-) indicates its absence.}
\begin{tabular}{lrrrrrrrr}
\hline
\textbf{Metric} & \textbf{F- C- D-} & \textbf{F+ C- D-} & \textbf{F- C+ D-} & \textbf{F- C- D+} & \textbf{F+ C+ D+} & \textbf{WangchanX} & \textbf{Typhoon} & \textbf{OpenThai} \\
\midrule
BERTScore & 53.22 & 62.66 & 52.55 & 52.28 & \underline{73.55} & 73.36 & \textbf{78.52} & 66.44 \\
BLEU & 0.00 & 0.00 & 0.00 & 0.00 & 0.01 & \textbf{0.85} & 0.16 & \underline{0.31} \\
ChrF & 4.62 & 3.85 & 2.41 & 2.74 & 7.68 & \textbf{14.88} & 11.76 & \underline{12.07} \\
METEOR & 2.33 & 2.96 & 1.33 & 1.70 & 5.32 & \textbf{8.72} & \underline{8.25} & 6.60 \\
ROUGE-1 & 1.05 & 2.46 & 1.34 & 1.03 & 4.58 & \underline{5.70} & \textbf{7.57} & 3.78 \\
ROUGE-2 & 0.07 & 0.59 & 0.04 & 0.19 & \underline{1.59} & 1.14 & \textbf{2.37} & 1.00 \\
ROUGE-L & 1.01 & 2.33 & 1.33 & 1.01 & 4.45 & \underline{5.16} & \textbf{7.35} & 3.50 \\
ROUGE-Lsum & 0.99 & 2.32 & 1.33 & 1.00 & 4.43 & \underline{5.21} & \textbf{7.36} & 3.51 \\
SQuAD F1 & 0.52 & 2.49 & 0.64 & 0.95 & 3.47 & \textbf{3.96} & \underline{3.51} & 2.04 \\
\hline
\multicolumn{9}{l}{\textit{(Thai Culture Test Set)}}
\vspace{0.5em}
\end{tabular}

\begin{tabular}{lrrrrrrrr}
\hline
\textbf{Metric} & \textbf{F- C- D-} & \textbf{F+ C- D-} & \textbf{F- C+ D-} & \textbf{F- C- D+} & \textbf{F+ C+ D+} & \textbf{WangchanX} & \textbf{Typhoon} & \textbf{OpenThai} \\
\midrule
BERTScore & 60.33 & 64.61 & 56.39 & 54.93 & 75.55 & \underline{75.70} & \textbf{80.03} & 71.11 \\
BLEU & 0.01 & 0.00 & 0.00 & 0.00 & 0.01 & \underline{0.33} & 0.16 & \textbf{0.61} \\
ChrF & 5.19 & 3.40 & 2.75 & 2.60 & 8.03 & \underline{12.52} & 11.55 & \textbf{13.78} \\
METEOR & 2.78 & 2.78 & 1.65 & 1.70 & 5.55 & 7.52 & \textbf{8.18} & \underline{8.11} \\
ROUGE-1 & 1.29 & 2.35 & 0.61 & 1.27 & 4.88 & 6.19 & \textbf{8.12} & \underline{6.48} \\
ROUGE-2 & 0.20 & 0.72 & 0.07 & 0.31 & 1.38 & \underline{2.30} & \textbf{3.19} & 2.21 \\
ROUGE-L & 1.29 & 2.33 & 0.60 & 1.24 & 4.88 & 5.96 & \textbf{7.96} & \underline{6.40} \\
ROUGE-Lsum & 1.26 & 2.32 & 0.61 & 1.27 & 4.90 & 5.93 & \textbf{7.98} & \underline{6.35} \\
SQuAD F1 & 1.01 & 2.46 & 0.96 & 1.11 & 3.03 & \underline{3.67} & \textbf{3.81} & 3.30 \\
\hline
\multicolumn{9}{l}{\textit{(General Test Set)}}
\end{tabular}
\label{tab:classification_results}
\end{table*}

\begin{table*}[ht!]
\centering
\small
\caption{Average evaluation results for the Closed QA task on the Thai Culture and General Test Sets. F, C, and D denote Fluency, Culture, and Diversity, respectively. The plus sign (+) indicates the presence of the corresponding attribute, while the minus sign (-) indicates its absence.}
\begin{tabular}{lrrrrrrrr}
\hline
\textbf{Metric} & \textbf{F- C- D-} & \textbf{F+ C- D-} & \textbf{F- C+ D-} & \textbf{F- C- D+} & \textbf{F+ C+ D+} & \textbf{WangchanX} & \textbf{Typhoon} & \textbf{OpenThai} \\
\midrule
BERTScore & 18.18 & 48.22 & 42.27 & 36.98 & 60.91 & \underline{61.49} & \textbf{65.67} & 44.32 \\
BLEU & 0.01 & 0.00 & 0.00 & 0.00 & 0.00 & \textbf{1.20} & \underline{1.09} & 0.00 \\
ChrF & 1.97 & 3.67 & 2.08 & 1.25 & 7.51 & \underline{14.62} & \textbf{15.12} & 4.40 \\
METEOR & 0.98 & 2.44 & 1.35 & 0.98 & 6.91 & \underline{12.52} & \textbf{13.89} & 2.92 \\
ROUGE-1 & 0.42 & 2.87 & 1.39 & 0.63 & 10.94 & \underline{16.86} & \textbf{19.74} & 5.10 \\
ROUGE-2 & 0.29 & 0.68 & 0.17 & 0.00 & 5.99 & \underline{10.45} & \textbf{12.50} & 2.02 \\
ROUGE-L & 0.42 & 2.80 & 1.36 & 0.62 & 10.76 & \underline{16.07} & \textbf{19.14} & 4.87 \\
ROUGE-Lsum & 0.42 & 2.82 & 1.36 & 0.64 & 10.86 & \underline{16.07} & \textbf{19.07} & 4.87 \\
SQuAD F1 & 0.29 & 2.23 & 0.83 & 0.23 & 9.01 & \textbf{14.46} & \underline{13.31} & 2.00 \\
\hline
\multicolumn{9}{l}{\textit{(Thai Culture Test Set)}}
\vspace{0.5em}
\end{tabular}

\begin{tabular}{lrrrrrrrr}
\hline
\textbf{Metric} & \textbf{F- C- D-} & \textbf{F+ C- D-} & \textbf{F- C+ D-} & \textbf{F- C- D+} & \textbf{F+ C+ D+} & \textbf{WangchanX} & \textbf{Typhoon} & \textbf{OpenThai} \\
\midrule
BERTScore & 32.35 & 56.25 & 50.62 & 40.48 & 68.44 & \underline{70.43} & \textbf{77.57} & 56.49 \\
BLEU & 0.11 & 0.00 & 0.00 & 0.00 & 0.00 & \textbf{4.20} & \underline{3.68} & 0.00 \\
ChrF & 4.16 & 5.20 & 2.70 & 0.95 & 7.98 & \underline{19.93} & \textbf{20.49} & 5.45 \\
METEOR & 1.91 & 3.43 & 1.50 & 0.61 & 5.94 & \underline{15.15} & \textbf{15.92} & 3.13 \\
ROUGE-1 & 1.12 & 3.62 & 1.34 & 0.32 & 8.25 & \underline{16.99} & \textbf{21.40} & 4.86 \\
ROUGE-2 & 0.37 & 1.89 & 0.33 & 0.05 & 5.04 & \underline{11.03} & \textbf{13.40} & 2.24 \\
ROUGE-L & 1.15 & 3.57 & 1.32 & 0.24 & 8.21 & \underline{16.30} & \textbf{20.44} & 4.75 \\
ROUGE-Lsum & 1.12 & 3.59 & 1.31 & 0.24 & 8.14 & \underline{16.33} & \textbf{20.47} & 4.70 \\
SQuAD F1 & 0.65 & 2.85 & 0.58 & 0.25 & 7.88 & \textbf{16.06} & \underline{14.00} & 1.75 \\
\hline
\multicolumn{9}{l}{\textit{(General Test Set)}}
\end{tabular}
\label{tab:closedqa_results}
\end{table*}

\begin{table*}[ht!]
\centering
\small
\caption{Average evaluation results for the Creative Writing task on the Thai Culture and General Test Sets. F, C, and D denote Fluency, Culture, and Diversity, respectively. The plus sign (+) indicates the presence of the corresponding attribute, while the minus sign (-) indicates its absence.}
\begin{tabular}{lrrrrrrrr}
\hline
\textbf{Metric} & \textbf{F- C- D-} & \textbf{F+ C- D-} & \textbf{F- C+ D-} & \textbf{F- C- D+} & \textbf{F+ C+ D+} & \textbf{WangchanX} & \textbf{Typhoon} & \textbf{OpenThai} \\
\midrule
BERTScore & 52.10 & 52.74 & 47.08 & 48.56 & 64.13 & 64.18 & \textbf{79.61} & \underline{69.70} \\
BLEU & 0.00 & 0.00 & 0.00 & 0.01 & 0.01 & \underline{0.49} & \textbf{1.38} & 0.32 \\
ChrF & 3.79 & 4.25 & 2.14 & 3.74 & 8.18 & \underline{18.98} & \textbf{20.35} & 15.52 \\
METEOR & 1.94 & 2.05 & 1.23 & 2.13 & 3.68 & \underline{9.06} & \textbf{12.81} & 8.28 \\
ROUGE-1 & 1.02 & 2.42 & 0.31 & 1.51 & 4.93 & 9.71 & \textbf{34.92} & \underline{17.03} \\
ROUGE-2 & 0.00 & 1.05 & 0.00 & 0.31 & 2.50 & 2.78 & \textbf{26.55} & \underline{10.03} \\
ROUGE-L & 0.99 & 2.42 & 0.31 & 1.42 & 4.60 & 8.85 & \textbf{35.16} & \underline{16.82} \\
ROUGE-Lsum & 0.99 & 2.42 & 0.31 & 1.46 & 4.65 & 9.05 & \textbf{34.47} & \underline{16.81} \\
\hline
\multicolumn{9}{l}{\textit{(Thai Culture Test Set)}}
\vspace{0.5em}
\end{tabular}

\begin{tabular}{lrrrrrrrr}
\hline
\textbf{Metric} & \textbf{F- C- D-} & \textbf{F+ C- D-} & \textbf{F- C+ D-} & \textbf{F- C- D+} & \textbf{F+ C+ D+} & \textbf{WangchanX} & \textbf{Typhoon} & \textbf{OpenThai} \\
\midrule
BERTScore & 52.34 & 54.56 & 47.36 & 50.82 & \underline{69.37} & 69.33 & \textbf{77.02} & 69.20 \\
BLEU & 0.00 & 0.00 & 0.00 & 0.00 & 0.02 & \underline{1.15} & \textbf{1.46} & 1.08 \\
ChrF & 4.30 & 4.16 & 2.07 & 4.52 & 9.03 & \textbf{21.93} & 20.77 & \underline{21.87} \\
METEOR & 2.40 & 2.40 & 1.14 & 2.43 & 5.06 & \underline{11.41} & \textbf{12.47} & 11.03 \\
ROUGE-1 & 2.11 & 2.61 & 0.34 & 1.50 & 8.12 & \underline{15.68} & \textbf{22.92} & 13.56 \\
ROUGE-2 & 1.00 & 0.83 & 0.00 & 0.35 & 3.37 & \underline{8.76} & \textbf{14.07} & 6.55 \\
ROUGE-L & 1.79 & 2.66 & 0.34 & 1.47 & 8.00 & \underline{15.18} & \textbf{22.17} & 12.92 \\
ROUGE-Lsum & 1.80 & 2.68 & 0.34 & 1.48 & 8.05 & \underline{15.37} & \textbf{22.07} & 12.99 \\
\hline
\multicolumn{9}{l}{\textit{(General Test Set)}}
\end{tabular}
\label{tab:creativewriting_results}
\end{table*}

\begin{table*}[ht!]
\centering
\small
\caption{Average evaluation results for the Multiple Choice task on the Thai Culture and General Test Sets. F, C, and D denote Fluency, Culture, and Diversity, respectively. The plus sign (+) indicates the presence of the corresponding attribute, while the minus sign (-) indicates its absence.}
\begin{tabular}{lrrrrrrrr}
\hline
\textbf{Metric} & \textbf{F- C- D-} & \textbf{F+ C- D-} & \textbf{F- C+ D-} & \textbf{F- C- D+} & \textbf{F+ C+ D+} & \textbf{WangchanX} & \textbf{Typhoon} & \textbf{OpenThai} \\
\midrule
BERTScore & 53.65 & 66.32 & 52.86 & 47.64 & \underline{70.36} & 69.91 & 53.20 & \textbf{71.94} \\
BLEU & 0.00 & 0.04 & 0.00 & 0.00 & 0.04 & \textbf{2.65} & 0.00 & \underline{2.30} \\
ChrF & 5.00 & 8.76 & 3.29 & 1.81 & 9.89 & \underline{15.95} & 5.54 & \textbf{18.79} \\
METEOR & 2.76 & 7.00 & 1.91 & 1.51 & 7.74 & \underline{11.61} & 4.62 & \textbf{11.98} \\
ROUGE-1 & 2.10 & 8.13 & 4.51 & 3.99 & 10.51 & \textbf{17.70} & 15.57 & \underline{17.56} \\
ROUGE-2 & 0.22 & 1.64 & 0.29 & 0.32 & 2.82 & \textbf{4.65} & 3.02 & \underline{4.31} \\
ROUGE-L & 2.13 & 8.03 & 4.30 & 3.87 & 9.77 & \textbf{16.71} & 14.79 & \underline{16.02} \\
ROUGE-Lsum & 2.08 & 8.12 & 4.29 & 3.95 & 9.89 & \textbf{16.79} & 14.89 & \underline{16.05} \\
SQuAD F1 & 0.73 & 5.12 & 1.38 & 1.41 & 6.39 & \textbf{10.97} & 7.18 & \underline{7.21} \\
\hline
\multicolumn{9}{l}{\textit{(Thai Culture Test Set)}}
\vspace{0.5em}
\end{tabular}

\begin{tabular}{lrrrrrrrr}
\hline
\textbf{Metric} & \textbf{F- C- D-} & \textbf{F+ C- D-} & \textbf{F- C+ D-} & \textbf{F- C- D+} & \textbf{F+ C+ D+} & \textbf{WangchanX} & \textbf{Typhoon} & \textbf{OpenThai} \\
\midrule
BERTScore & 56.41 & 65.37 & 54.17 & 42.19 & \textbf{70.90} & 66.83 & 54.80 & \underline{69.97} \\
BLEU & 0.01 & 0.03 & 0.00 & 0.00 & 0.06 & \underline{0.37} & 0.00 & \textbf{1.70} \\
ChrF & 5.03 & 7.86 & 3.23 & 0.95 & 9.60 & \underline{11.00} & 4.44 & \textbf{15.02} \\
METEOR & 2.65 & 5.78 & 1.99 & 0.93 & 7.40 & \underline{7.99} & 4.45 & \textbf{9.40} \\
ROUGE-1 & 1.46 & 3.82 & 3.93 & 3.34 & 6.69 & \underline{12.34} & \textbf{13.34} & 10.10 \\
ROUGE-2 & 0.23 & 0.93 & 0.16 & 0.44 & 1.76 & \textbf{3.39} & \underline{2.23} & 1.66 \\
ROUGE-L & 1.39 & 3.70 & 3.96 & 3.28 & 6.29 & \underline{11.44} & \textbf{13.40} & 9.39 \\
ROUGE-Lsum & 1.41 & 3.75 & 3.97 & 3.33 & 6.32 & \underline{11.43} & \textbf{13.38} & 9.39 \\
SQuAD F1 & 0.84 & 2.98 & 1.09 & 0.86 & 5.05 & \textbf{6.63} & \underline{6.60} & 4.22 \\
\hline
\multicolumn{9}{l}{\textit{(General Test Set)}}
\end{tabular}
\label{tab:multiplechoice_results}
\end{table*}

\begin{table*}[ht!]
\centering
\small
\caption{Average evaluation results for the Open QA task on the Thai Culture and General Test Sets. F, C, and D denote Fluency, Culture, and Diversity, respectively. The plus sign (+) indicates the presence of the corresponding attribute, while the minus sign (-) indicates its absence.}
\begin{tabular}{lrrrrrrrr}
\hline
\textbf{Metric} & \textbf{F- C- D-} & \textbf{F+ C- D-} & \textbf{F- C+ D-} & \textbf{F- C- D+} & \textbf{F+ C+ D+} & \textbf{WangchanX} & \textbf{Typhoon} & \textbf{OpenThai} \\
\midrule
BERTScore & 52.40 & 58.74 & 49.79 & 50.44 & \underline{71.47} & 68.23 & \textbf{76.95} & 68.09 \\
BLEU & 0.00 & 0.00 & 0.00 & 0.00 & 0.02 & \textbf{0.82} & 0.70 & \underline{0.75} \\
ChrF & 4.17 & 4.17 & 2.26 & 2.97 & 8.27 & 14.64 & \underline{15.30} & \textbf{15.77} \\
METEOR & 2.08 & 2.65 & 1.22 & 1.73 & 5.22 & 7.56 & \textbf{9.30} & \underline{7.98} \\
ROUGE-1 & 1.03 & 2.57 & 1.04 & 2.22 & 6.26 & 10.30 & \textbf{12.64} & \underline{11.19} \\
ROUGE-2 & 0.04 & 0.69 & 0.15 & 0.45 & 2.47 & 4.60 & \textbf{6.74} & \underline{5.73} \\
ROUGE-L & 0.91 & 2.53 & 1.03 & 2.22 & 6.12 & 9.81 & \textbf{12.29} & \underline{10.77} \\
ROUGE-Lsum & 0.91 & 2.52 & 1.02 & 2.24 & 6.14 & 9.81 & \textbf{12.29} & \underline{10.77} \\
SQuAD F1 & 0.46 & 1.29 & 0.44 & 0.68 & 2.41 & \underline{3.80} & \textbf{4.37} & 3.07 \\
\hline
\multicolumn{9}{l}{\textit{(Thai Culture Test Set)}}
\vspace{0.5em}
\end{tabular}

\begin{tabular}{lrrrrrrrr}
\hline
\textbf{Metric} & \textbf{F- C- D-} & \textbf{F+ C- D-} & \textbf{F- C+ D-} & \textbf{F- C- D+} & \textbf{F+ C+ D+} & \textbf{WangchanX} & \textbf{Typhoon} & \textbf{OpenThai} \\
\midrule
BERTScore & 58.49 & 60.27 & 51.89 & 53.20 & \underline{75.97} & 72.78 & \textbf{80.28} & 74.20 \\
BLEU & 0.00 & 0.00 & 0.00 & 0.00 & 0.04 & 0.46 & \underline{0.71} & \textbf{1.22} \\
ChrF & 4.69 & 3.93 & 2.33 & 3.08 & 9.18 & 14.12 & \underline{15.18} & \textbf{18.10} \\
METEOR & 2.39 & 2.39 & 1.26 & 1.72 & 5.63 & 7.56 & \underline{9.52} & \textbf{9.58} \\
ROUGE-1 & 1.52 & 1.63 & 1.19 & 1.46 & 3.89 & 9.22 & \underline{12.38} & \textbf{13.74} \\
ROUGE-2 & 0.27 & 0.42 & 0.07 & 0.35 & 1.37 & 3.85 & \underline{6.18} & \textbf{7.09} \\
ROUGE-L & 1.46 & 1.62 & 1.20 & 1.45 & 3.86 & 8.75 & \underline{11.83} & \textbf{13.16} \\
ROUGE-Lsum & 1.45 & 1.63 & 1.21 & 1.44 & 3.85 & 8.73 & \underline{11.87} & \textbf{13.21} \\
SQuAD F1 & 0.93 & 0.87 & 0.68 & 0.70 & 2.37 & 3.35 & \textbf{4.61} & \underline{3.95} \\
\hline
\multicolumn{9}{l}{\textit{(General Test Set)}}
\end{tabular}
\label{tab:openqa_results}
\end{table*}

\begin{table*}[ht!]
\centering
\small
\caption{Average evaluation results for the Summarization task on the Thai Culture and General Test Sets. F, C, and D denote Fluency, Culture, and Diversity, respectively. The plus sign (+) indicates the presence of the corresponding attribute, while the minus sign (-) indicates its absence.}
\begin{tabular}{lrrrrrrrr}
\hline
\textbf{Metric} & \textbf{F- C- D-} & \textbf{F+ C- D-} & \textbf{F- C+ D-} & \textbf{F- C- D+} & \textbf{F+ C+ D+} & \textbf{WangchanX} & \textbf{Typhoon} & \textbf{OpenThai} \\
\midrule
BERTScore & 37.11 & 57.15 & 51.25 & 47.76 & \underline{75.77} & 75.74 & \textbf{84.06} & 59.31 \\
BLEU & 0.11 & 0.03 & 0.00 & 0.00 & 0.57 & \underline{9.25} & \textbf{11.71} & 0.82 \\
ChrF & 6.08 & 7.60 & 5.08 & 3.12 & 15.19 & \underline{27.36} & \textbf{33.98} & 12.82 \\
METEOR & 2.88 & 6.47 & 3.20 & 2.33 & 12.67 & \underline{21.91} & \textbf{28.89} & 8.07 \\
ROUGE-1 & 0.42 & 5.00 & 2.52 & 2.93 & 10.97 & \underline{21.10} & \textbf{32.22} & 8.25 \\
ROUGE-2 & 0.09 & 2.06 & 0.52 & 0.95 & 5.19 & \underline{12.48} & \textbf{19.73} & 4.06 \\
ROUGE-L & 0.42 & 4.80 & 2.47 & 2.88 & 10.80 & \underline{20.27} & \textbf{30.71} & 7.93 \\
ROUGE-Lsum & 0.41 & 4.83 & 2.48 & 2.89 & 10.82 & \underline{20.33} & \textbf{30.72} & 7.95 \\
\hline
\multicolumn{9}{l}{\textit{(Thai Culture Test Set)}}
\vspace{0.5em}
\end{tabular}

\begin{tabular}{lrrrrrrrr}
\hline
\textbf{Metric} & \textbf{F- C- D-} & \textbf{F+ C- D-} & \textbf{F- C+ D-} & \textbf{F- C- D+} & \textbf{F+ C+ D+} & \textbf{WangchanX} & \textbf{Typhoon} & \textbf{OpenThai} \\
\midrule
BERTScore & 40.50 & 58.27 & 51.96 & 46.54 & \underline{78.67} & 78.37 & \textbf{85.97} & 59.96 \\
BLEU & 0.16 & 0.01 & 0.00 & 0.00 & 0.37 & \underline{7.55} & \textbf{11.14} & 0.18 \\
ChrF & 5.89 & 6.48 & 3.53 & 1.88 & 13.82 & \underline{24.81} & \textbf{32.39} & 9.78 \\
METEOR & 3.02 & 5.19 & 2.43 & 1.50 & 11.72 & \underline{19.82} & \textbf{28.37} & 6.43 \\
ROUGE-1 & 1.32 & 5.23 & 2.10 & 1.34 & 9.55 & \underline{21.80} & \textbf{32.27} & 6.35 \\
ROUGE-2 & 0.24 & 2.23 & 0.31 & 0.36 & 4.35 & \underline{13.03} & \textbf{19.76} & 2.93 \\
ROUGE-L & 1.27 & 5.09 & 2.06 & 1.29 & 9.37 & \underline{20.72} & \textbf{30.52} & 6.04 \\
ROUGE-Lsum & 1.27 & 5.09 & 2.06 & 1.29 & 9.34 & \underline{20.74} & \textbf{30.53} & 6.05 \\
\hline
\multicolumn{9}{l}{\textit{(General Test Set)}}
\end{tabular}
\label{tab:summarization_results}
\end{table*}